\documentclass[11pt]{article}

\usepackage[final]{acl}

\usepackage{microtype}
\usepackage{graphicx}
\usepackage{booktabs}
\usepackage{array}
\usepackage{amsmath}
\usepackage{amssymb}
\usepackage{subcaption}
\usepackage{tikz}
\usetikzlibrary{arrows.meta,positioning,calc}

\usepackage{fontspec}
\newfontfamily\tamilfont[
  Path        = ./,
  Extension   = .ttf,
  UprightFont = *-Regular,
  BoldFont    = *-Bold,
  ItalicFont  = *-Italic,
  Script      = Tamil,
  Scale       = 0.95]{NotoSerifTamil}

\newcommand{\ta}[1]{{\tamilfont #1}}

\title{Contextual Tamil Spelling and Grammar Correction Using\\
Progressively Fine-Tuned Sequence-to-Sequence Transformers}

\author{
  Karthikeyan A\textsuperscript{1},
  Jaya Nirmala S\textsuperscript{1},
  Sangeetha Sivanesan\textsuperscript{1},
  Indhu R\textsuperscript{2}, \\
  \textbf{Pranav Kumar\textsuperscript{1},
  Bharat Jude Johnson\textsuperscript{1},
  Vishnu Ram\textsuperscript{1}} \\[2pt]
  \textsuperscript{1}National Institute of Technology, Tiruchirappalli \\
  \textsuperscript{2}Tamil University, Thanjavur \\
  \textsuperscript{1}Email: \href{mailto:vsga0026@gmail.com}{vsga0026@gmail.com}
}

\begin{document}
\maketitle

\begin{abstract}
Tamil spell and grammar correction is challenging because Tamil is an
agglutinative low-resource language with rich verbal morphology, complex
sandhi (phonetic transformation) rules at word boundaries, and a script of
247 distinct letters. Prior work targets word-level surface errors with
rule-based methods, statistical $n$-gram models, Minimum Edit Distance, or
hybrid pipelines with a transformer re-ranker; such methods cannot reliably
handle contextual errors --- subject--verb agreement, tense consistency, or
cross-word sandhi --- which require sentence-level understanding. We propose
an end-to-end sequence-to-sequence formulation and fine-tune mT5-small and
mBART-50 on a synthetic corpus of up to 657,720 noisy--clean Tamil sentence
pairs spanning ten error categories. Both backbones follow the same
four-stage progressive schedule, each stage targeting one weakness: surface
noise (v2), contextual grammar (v3), single-site sandhi (v4), and multi-site
cross-word sandhi (v5). On a 1,000-sentence balanced diagnostic set verified
disjoint from all training data, our best model, mBART-50 v5, reaches 69.3\%
top-1 exact-match accuracy, with 87.5\% on sandhi and 43.5\% on
subject--verb agreement. The schedule is what produces these gains:
subject--verb accuracy rises from 1.0\% to 52.5\% once contextual pairs are
introduced, and sandhi from 0\% to 87.5\% once multi-site sandhi pairs are.
We additionally quantify a precision--recall trade-off this literature has
not reported: sandhi recall is paid for monotonically in identity accuracy.
Finally, Tamil-LLaMA-7B-Instruct reaches 19.0\% zero-shot and 24.7\% with
three demonstrations against a 20.0\% copy baseline, showing that a
Tamil-adapted instruction model does not transfer to specialised
sentence-level correction without task-specific supervision.
\end{abstract}

\section{Introduction}

Tamil is one of the oldest classical languages in the world, spoken by over
75 million people, with a literary tradition spanning more than 2,000 years.
It uses 12 vowels (\emph{uyir ezhuthukal}), 18 consonants (\emph{mei
ezhuthukal}), and 216 compound characters (\emph{uyirmei ezhuthukal}) --- a
total of 247 distinct letters. Its agglutinative morphology packs tense,
person, gender, number and honorific information into single verb endings,
which means that spelling and grammatical correctness are deeply intertwined
and context-dependent.

While spell correction has been studied extensively for English and Hindi,
Tamil has received comparatively limited attention. In English, errors are
largely surface-level and can be corrected by edit-distance models trained
on large datasets. In Tamil, errors extend far beyond character-level
mistakes. The words \ta{பழம்} (``fruit'') and \ta{பலம்} (``strength'')
differ by one letter but mean entirely different things. Tamil also follows
strict sandhi rules: the phrase \ta{அதை கொண்டு வா} (``bring that'') should
be written \ta{அதைக் கொண்டு வா}, with the consonant \ta{க்} inserted because
of the \emph{vallinam}-triggered transformation. The agglutinative nature of
Tamil and the low-resource setting together make Tamil spell correction
substantially harder than for high-resource languages.

Existing Tamil spell-correction approaches struggle along three dimensions.
First, they typically address only a narrow set of error types (usually
surface-level phonetic substitution) and ignore contextual errors. Second,
they often produce a ranked list of candidates rather than a single
correction, which is unsuitable for real-time applications. Third, recent
hybrid work that combines transformer re-ranking with statistical retrieval
treats the transformer only as a scoring function, leaving its generative
capacity unused. None of these approaches reliably handles subject--verb
agreement, tense consistency or cross-word sandhi at the sentence level.

To address these challenges, this work:
\begin{enumerate}
\item an end-to-end sequence-to-sequence formulation that fine-tunes
  mT5-small and mBART-50 to map noisy Tamil sentences directly to clean
  ones, rather than using transformers as candidate re-rankers;
\item the same four-stage progressive schedule (v2 $\rightarrow$ v5) applied
  to both backbones, separating the contribution of curriculum from that of
  architecture;
\item a noise-generation pipeline producing 657,720 training pairs over ten
  error categories, including corpus-mined agreement errors and multi-site
  cross-word sandhi violations, with a 70/20/10 split per stage;
\item evaluation on a 1,000-sentence balanced diagnostic set verified
  disjoint from training data, reporting top-1 exact-match accuracy rather
  than the top-$k$ accuracy used in prior context-blind work; and
\item a controlled comparison against Tamil-LLaMA
  \citep{balachandran2023tamilllama} under prompt selection, and against a
  trivial copy baseline that establishes the floor.
\end{enumerate}

\begin{figure*}[t]
\centering
\tikzset{
  fbox/.style   = {draw, rounded corners=2pt, align=center, inner sep=2.5pt,
                   font=\scriptsize, text width=2.45cm, minimum height=0.62cm},
  sbox/.style   = {draw, rounded corners=2pt, align=center, inner sep=2pt,
                   font=\tiny, text width=1.35cm, minimum height=0.62cm},
  arr/.style    = {-{Latex[length=1.4mm,width=1.1mm]}, shorten >=1pt, shorten <=1pt},
  lab/.style    = {font=\tiny, inner sep=1.2pt},
}
\begin{subfigure}[t]{0.345\textwidth}
\centering
\begin{tikzpicture}[node distance=4.5mm]
  \node[fbox] (wiki) {Tamil Wikipedia\\($\approx$17M words)};
  \node[fbox, below=of wiki] (clean) {Cleaning, dedup.,\\length filter (3--50)};
  \node[fbox, below=of clean] (sent) {1.2M clean\\Tamil sentences};
  \node[sbox] (g2) at ($(sent.south)+(0,-8mm)$) {Contextual-error generator};
  \node[sbox] (g1) at ($(g2)+(-1.75cm,0)$) {Surface-noise generator};
  \node[sbox] (g3) at ($(g2)+(1.75cm,0)$) {Sandhi-violation generator};
  \node[sbox, below=5mm of g1] (o1) {500,000 pairs,\\5 surface types};
  \node[sbox, below=5mm of g2] (o2) {575,000 pairs, 10 context.\ types};
  \node[sbox, below=5mm of g3] (o3) {30k single-site,\\40k multi-site};
  \draw[arr] (wiki) -- (clean);
  \draw[arr] (clean) -- (sent);
  \draw[arr] (sent.south) -- ++(0,-3mm) -| (g1.north);
  \draw[arr] (sent.south) -- ++(0,-3mm) -| (g2.north);
  \draw[arr] (sent.south) -- ++(0,-3mm) -| (g3.north);
  \draw[arr] (g1) -- (o1);
  \draw[arr] (g2) -- (o2);
  \draw[arr] (g3) -- (o3);
\end{tikzpicture}
\caption{Dataset construction}
\label{fig:arch-data}
\end{subfigure}
\hfill
\begin{subfigure}[t]{0.325\textwidth}
\centering
\begin{tikzpicture}[node distance=5.2mm]
  \node[fbox] (bb) {Pre-trained backbone\\mT5-small / mBART-50};
  \node[fbox, below=of bb] (s1) {Stage 1 (v2)\\surface noise};
  \node[fbox, below=of s1] (s2) {Stage 2 (v3)\\contextual grammar};
  \node[fbox, below=of s2] (s3) {Stage 3 (v4)\\single-site sandhi};
  \node[fbox, below=of s3] (s4) {Stage 4 (v5)\\multi-site sandhi};
  \node[fbox, below=of s4] (fin) {Final checkpoint\\mBART-50 v5};
  \draw[arr] (bb) -- node[lab, right] {500k surface} (s1);
  \draw[arr] (s1) -- node[lab, right] {575k context $+$82.7k aug.} (s2);
  \draw[arr] (s2) -- node[lab, right] {30k sandhi $+$30k replay} (s3);
  \draw[arr] (s3) -- node[lab, right] {40k multi-site $+$20k replay} (s4);
  \draw[arr] (s4) -- (fin);
  \draw[arr, dashed] (s2.west) -- ++(-3.5mm,0) |- node[lab, left, pos=0.25] {resume from v3} (s4.west);
\end{tikzpicture}
\caption{Progressive fine-tuning}
\label{fig:arch-train}
\end{subfigure}
\hfill
\begin{subfigure}[t]{0.29\textwidth}
\centering
\begin{tikzpicture}[node distance=5.2mm]
  \node[fbox] (in) {Noisy Tamil\\sentence};
  \node[fbox, below=of in] (norm) {Unicode normalisation\\(NFC, punctuation)};
  \node[fbox, below=of norm] (tok) {SentencePiece tokenizer\\\texttt{ta\_IN} language tag};
  \node[fbox, below=of tok] (enc) {Encoder\\12 layers};
  \node[fbox, below=of enc] (dec) {Decoder\\beam $=4$, no-repeat 3};
  \node[fbox, below=of dec] (out) {Corrected\\Tamil sentence};
  \draw[arr] (in) -- (norm);
  \draw[arr] (norm) -- (tok);
  \draw[arr] (tok) -- (enc);
  \draw[arr] (enc) -- (dec);
  \draw[arr] (dec) -- (out);
\end{tikzpicture}
\caption{Single-pass inference}
\label{fig:arch-inf}
\end{subfigure}
\caption{Overall system architecture. (a) Clean Tamil Wikipedia text passes
through three generators injecting surface, contextual and sandhi errors.
(b) One backbone is fine-tuned in four stages, each resuming from the
previous checkpoint and adding a new error class plus a replay sample
against catastrophic forgetting; v4 and v5 both resume from v3. (c) At
inference a sentence is corrected in one forward pass.}
\label{fig:architecture}
\end{figure*}
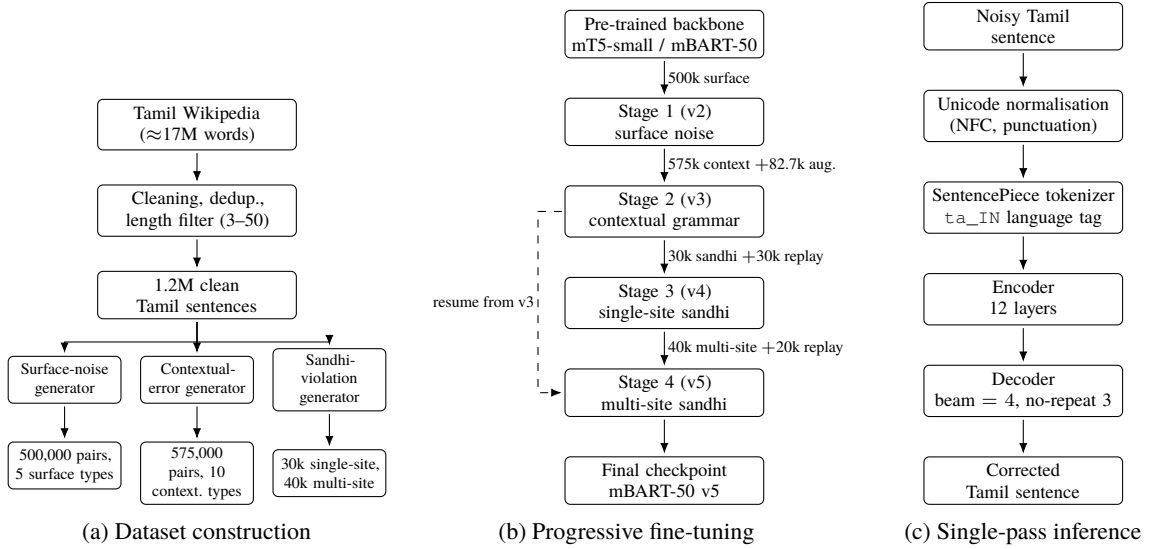

Figure~\ref{fig:architecture} gives the overall picture: dataset
construction, progressive fine-tuning, and single-pass inference.

% ---------------------------------------------------------------
% Figure 1
% ---------------------------------------------------------------

\section{Previous Work}

\paragraph{Rule-based and statistical approaches.}
\citet{parthasarathi2003tamil} proposed a morphological-analysis-based spell
checker using linguistic rules; the work reports no quantitative evaluation,
and rule-based systems in general struggle to scale to diverse error
patterns or unseen words. Solthiruthi \citep{elanjelian2004solthiruthi} and
Vaani \citep{rajaraman2014vaani} are publicly available tools in the same
category. \citet{segar2015contextual} proposed a bigram-based contextual
spell checker achieving 89.13\% accuracy; its two-word context window is the
main limitation, because Tamil grammatical dependencies frequently span more
than two words. \citet{sakuntharaj2016novel} introduced a hybrid tree-based
$n$-gram approach reporting 91\% accuracy on non-word errors only, and
\citet{sakuntharaj2018detecting} reported 98\% on real-word errors under the
strong assumption that every word is individually valid.
\citet{kumar2020design} used Minimum Edit Distance for Tamil correction.

\paragraph{Deep-learning and hybrid approaches.}
\citet{etoori2018automatic} introduced a sequence-to-sequence deep-learning
model for Hindi and Telugu spell correction, generating synthetic training
pairs to overcome low-resource constraints; they did not extend the approach
to Tamil and did not model Tamil-specific phenomena such as sandhi.
\citet{sampath2023hybrid} combined edit distance, Soundex matching,
rule-based correction and an LSTM scoring component to achieve 95.67\%
accuracy. More recent hybrid approaches integrate MED, $n$-gram
probabilities, FastText \citep{bojanowski2017enriching} embeddings and
pre-trained transformer re-rankers; while effective for ranking
surface-level candidates, such approaches do not generate corrections and
cannot fix grammatical errors that the candidate generator did not propose.
\citet{sharma2025higec} explore the closely related problem of Hindi
grammatical error correction in a low-resource setting, comparing
direct-noise injection, round-trip translation and neural error generation
--- evidence that synthetic-data strategies are the standard answer to the
absence of annotated Indic GEC corpora. Tamil-specific neural work includes
DDSpell \citep{uthayamoorthy2019ddspell}, a context-aware Sinhala--Tamil
correction environment \citep{sithamparanathan2019sinhala}, and a RoBERTa
spell checker \citep{rajalakshmi2023context}, all of which judge
pre-enumerated candidates rather than generate a corrected sentence.

\paragraph{Multilingual transformers for seq2seq correction.}
\citet{xue2021mt5} introduced mT5, a multilingual text-to-text transformer
pre-trained on the mC4 corpus across 101 languages including Tamil.
\citet{liu2020multilingual} and \citet{tang2020multilingual} developed mBART
and its 50-language extension, both pre-trained with a denoising
sequence-to-sequence objective conceptually close to spell correction.
\citet{elango2023tamil} fine-tune a multilingual T5 for Tamil error
correction, and \citet{yazhmozhi2026tamilmayangoli} compare mBART, mT5 and
NLLB \citep{nllb2024scaling} on one error class; to our knowledge none
applies these backbones to a taxonomy spanning surface, agreement and
cross-word sandhi errors under a single sentence-level model.

\paragraph{Mayangoli-specific correction.}
Closest to our setting is TamilMayangoliSpell
\citep{yazhmozhi2026tamilmayangoli}, which also fine-tunes multilingual
seq2seq models on synthetic Tamil pairs and reports 93.50\% exact match with
mT5. It targets a single class --- Mayangoli confusions among phonetically
similar graphemes (\ta{ல}/\ta{ள}/\ta{ழ}, \ta{ர}/\ta{ற},
\ta{ந}/\ta{ன}/\ta{ண}), corresponding to our phonetic category alone ---
with substitutions constrained to remain dictionary-valid, and explicitly
excludes sandhi, Kuril--Nedil and non-word errors. It evaluates on a 10\%
split of the same induced distribution used for training, with one error per
sentence and no no-edit items. Its pipeline, data and models are publicly
released, which our work does not yet match.

\paragraph{Tamil-adapted large language models.}
Tamil-LLaMA \citep{balachandran2023tamilllama} extends LLaMA-2-7B by adding
16K Tamil tokens, continually pre-training on Tamil text and applying
instruction tuning. Its performance on focused spell-correction tasks has
not previously been measured.

Compared with prior Tamil-specific work, our approach (i) uses end-to-end
seq2seq generation rather than retrieval-plus-ranking, (ii) covers ten error
categories including agreement and cross-word sandhi rather than a single
confusion class, (iii) reports top-1 accuracy on a balanced set verified
disjoint from training data, with a 200-item identity slice and a copy
baseline fixing the floor, and (iv) compares directly against a
Tamil-adapted LLM.

\section{Dataset Creation}

The shortage of large-scale annotated Tamil error corpora is the central
obstacle for deep-learning approaches to Tamil spell correction. Existing
resources do not capture the diversity of errors users actually make,
particularly contextual phenomena. We therefore construct a synthetic
dataset by injecting controlled noise into clean Tamil text. All training
and evaluation pairs in this work are synthetic in this sense; no corpus of
authentic annotated Tamil spelling errors is publicly available.

\subsection{Corpus}

We use the Tamil Wikipedia dataset \citep{gaurav2019tamilwiki} as our source
of clean Tamil text, comprising approximately 17 million words. The
pre-processing pipeline removes non-Tamil characters, English words, XML
tags, parenthetical content and punctuation. After deduplication and length
filtering (3 to 50 words per sentence) we obtain approximately 1.2 million
clean Tamil sentences, which form the basis for noise generation.
Encyclopaedic prose differs in register, sentence length and vocabulary from
the messaging, search and student-writing contexts in which a spell checker
is most used, so both the training corpus and the diagnostic set inherit a
Wikipedia domain bias. The Tamil Wikipedia content is used under the CC
BY-SA 4.0 licence, consistent with its intended use for research purposes.

\begin{table}[t]
\centering
\small
\begin{tabular}{lccc}
\toprule
Error type & Dist.\ (\%) & Right & Wrong \\
\midrule
Insertion            & 13--15 & \ta{தளிகை}  & \ta{தளிக்கை} \\
Deletion             & 7--9   & \ta{அவனைச்} & \ta{அவனை}    \\
Subst.\ (cons.)      & 27--29 & \ta{மனம்}   & \ta{மநம்}     \\
Subst.\ (mathirai)   & 21--23 & \ta{மலை}    & \ta{மல}      \\
Transposition        & 25--27 & \ta{பள்ளம்} & \ta{பளள்ம்}   \\
\bottomrule
\end{tabular}
\caption{Surface-level error categories in the base 500,000-pair
surface-noise dataset.}
\label{tab:surface}
\end{table}

\begin{table}[t]
\centering
\small
\begin{tabular}{lr}
\toprule
Category & Proportion (\%) \\
\midrule
Subject--verb (gender)        & 26.5 \\
Tense consistency             & 15.1 \\
Case marker                   & 10.4 \\
Plural agreement              & 8.5 \\
Honorific agreement           & 8.3 \\
Compound verb splitting       & 6.3 \\
Verb form (feminine subject)  & 6.3 \\
Negative verb                 & 6.2 \\
Redundant connector           & 4.0 \\
Corpus-mined verb swap        & 2.3 \\
Identity (no-change pairs)    & 16.5 \\
\bottomrule
\end{tabular}
\caption{Contextual error categories in the 75,000-pair contextual
augmentation set.}
\label{tab:contextual}
\end{table}

\subsection{Error Categories}

Our error taxonomy comprises ten categories designed to cover the range of
mistakes Tamil writers commonly produce. The categories fall into two
groups: surface-level errors that can be detected without sentence context,
and contextual errors that require understanding of the surrounding words.
Table~\ref{tab:surface} lists the surface-level types used in the base
500,000-pair dataset; Table~\ref{tab:contextual} lists the contextual
categories introduced in the 75,000-pair augmentation set. Sandhi violations
are structurally distinct because they operate across word boundaries, and
are addressed by dedicated datasets generated separately for v4 and v5
(Section~\ref{sec:sandhi-aug}).

\subsection{Noise Generation}

The dataset is built in three phases. The first produces 500,000
surface-noise pairs, injecting noise at the word level using the weighted
distribution in Table~\ref{tab:surface}, with single errors in 32--35\% of
words and double errors in 3--5\% (the per-word maximum). The second
generates 75,000 contextual pairs (Table~\ref{tab:contextual}), 45\% by
corpus mining --- substituting one verb ending in a real Wikipedia sentence
with an incorrect alternative of the same tense --- and the rest from
templates covering subject--verb combinations the corpus underrepresents.

\subsection{Sandhi-Specific Augmentation}
\label{sec:sandhi-aug}

Sandhi errors require special treatment because they cannot be modelled as
word-internal perturbations. We construct two sandhi datasets directly from
clean Wikipedia text. The first (30,000 pairs, used for v4) identifies
sentences containing one \emph{vallinam} sandhi site --- a word ending in
\ta{க்}, \ta{ச்}, \ta{த்} or \ta{ப்} followed by a word beginning with the
same consonant --- and removes exactly one suffix to produce the noisy
version. The second (40,000 pairs, used for v5) extends this by allowing up
to three sandhi sites to be removed per sentence: 32,425 pairs with one site
removed, 6,608 with two sites and 967 with three.

\subsection{Dataset Composition per Model}

Each stage in our progression is trained on a different combination of the
above components; Table~\ref{tab:composition} lists the composition. For v3
the augmentation block adds 2,720 subject--verb template pairs, 30,000
random phonetic perturbations, 20,000 within-word sandhi noise pairs and
30,000 identity pairs on top of the 575,000-pair main file, yielding 657,720
effective training pairs.

\subsection{Train, Validation and Test Splits}

Each stage corpus is split 70/20/10 into train, validation and test
portions. The split is hashed on the clean side of each pair, so a sentence
and all of its noisy variants always fall in the same partition and no
near-duplicate can straddle the boundary. The validation portion is used for
checkpoint selection by token-F1 during training; the 10\% test portion is
never seen during model selection. For the v3 stage, for example, this
yields 460,527 train, 131,112 validation and 66,081 test pairs.

\subsection{The Balanced Diagnostic Test Set}
\label{sec:diagnostic}

Aggregate accuracy on an in-distribution test slice hides which linguistic
phenomena a model has actually learned, so for final reporting we
additionally construct a balanced diagnostic set of 1,000 sentences: 200 per
category, covering phonetic confusion (\ta{ழ}/\ta{ள}/\ta{ல},
\ta{ற}/\ta{ர}, \ta{ண}/\ta{ன}), subject--verb agreement (gender, person,
number), pulli omission (missing dot \ta{◌்}), cross-word \emph{vallinam}
sandhi, and identity (correct sentences that must not be changed).

Crucially, the set is not assumed to be unseen --- it is verified to be.
Every candidate sentence is checked against all four training files by
exact-match hash and by 5-gram Jaccard overlap, and any candidate above the
overlap threshold is rejected and logged.

Because 200 of the 1,000 items are identity sentences that require no edit,
a trivial system that returns its input unchanged scores exactly 20.0\% by
construction. We report this copy baseline alongside all trained and
prompted systems, since it is the floor any correction system must clear.

\section{Proposed Work}

We propose a progressive fine-tuning pipeline applied to two multilingual
transformer architectures: mT5-small and mBART-50. The two models differ in
scale, pre-training objective and the way they signal the target language.
Applying the same four-stage schedule to both allows the effect of the
training curriculum to be separated from the effect of the architecture.

\subsection{Backbones}

\paragraph{mT5-small.}
mT5 \citep{xue2021mt5} is a multilingual text-to-text transformer
pre-trained with a span-corruption objective on mC4 across 101 languages. We
use the \texttt{mt5-small} checkpoint (300M parameters). Every input is
prefixed with the task instruction string \texttt{"correct tamil: "}
followed by the noisy Tamil sentence, and the model is trained to generate
the corrected sentence.

mT5's span-corruption pre-training introduced an implementation issue: the
model was trained to reconstruct masked spans using sentinel tokens
(\texttt{<extra\_id\_0>}, \ldots), and without intervention these tokens
leak into generated output. We override
\texttt{decoder\_start\_token\_id} to the pad token and remove
\texttt{forced\_bos\_token\_id}, so the decoder emits clean Tamil rather
than sentinel-prefixed sequences. We use Adafactor
\citep{shazeer2018adafactor} rather than AdamW for its lower memory
footprint on T5-family models and its stability under mixed precision, where
we observed NaN losses with AdamW.

\paragraph{mBART-50.}
mBART \citep{liu2020multilingual} and its 50-language extension
\citep{tang2020multilingual} are multilingual seq2seq models pre-trained
with a denoising objective that is conceptually very close to spell
correction: given a corrupted input, reconstruct the original. mBART-50 uses
explicit language-ID tokens (\texttt{ta\_IN} for Tamil) prepended to both
source and target sequences, which makes it particularly suitable for
monolingual transformations within a single language. We use the
\texttt{facebook/\allowbreak mbart-large-50-\allowbreak many-to-many-mmt} checkpoint (610M
parameters).

\subsection{The Progressive Fine-Tuning Schedule}

Rather than training a single model end-to-end on the union of all data, we
adopt a four-stage pipeline in which each stage begins from the best
checkpoint of the previous stage and targets a specific category of errors.
Progression decisions were made after qualitative inspection of error
patterns on the validation set and on the diagnostic set.

\paragraph{v2 --- surface noise.}
Fine-tunes the backbone on the 500,000-pair surface-noise dataset,
establishing baseline ability to handle insertion, deletion, substitution
and transposition errors.

\paragraph{v3 --- contextual augmentation.}
Building on v2, v3 incorporates the 75,000-pair contextual augmentation
(Table~\ref{tab:contextual}) plus 82,720 additional augmented pairs, for a
total of 657,720 training pairs. Training resumes from v2 at a lower
learning rate to prevent catastrophic forgetting of v2's surface abilities,
with a cosine schedule, 3\% warmup and early stopping (patience 3) on
token-F1.

\paragraph{v4 --- initial sandhi exposure.}
Error analysis of v3 showed that cross-word \emph{vallinam} sandhi was not
learned from the augmented in-word sandhi pairs alone. We generated a
30,000-pair dataset by removing one \emph{vallinam} sandhi suffix from real
corpus sentences and continued training from v3 for one epoch, with 30,000
replay samples from the v3 corpus.

\paragraph{v5 --- multi-site sandhi.}
To extend v4's single-site exposure to sentences containing multiple sandhi
sites, we constructed a 40,000-pair dataset removing 1, 2 or 3 sites per
sentence (32,425 / 6,608 / 967 respectively). Training resumed from v3 for 2
epochs with 20,000 replay samples. Table~\ref{tab:hyper} summarises the
hyperparameters per stage.

\begin{table*}[t]
\centering
\small
\begin{tabular}{lccccccc}
\toprule
Stage & Surface  & Contextual & Subject--verb & Phonetic & Sandhi & Identity & Total \\
      & noise (500k) & (575k) & templates & aug. & aug.\ (30k / 40k) & aug. & pairs \\
\midrule
v2 & \checkmark & --         & --    & --     & --                  & --                & 500,000 \\
v3 & \checkmark & \checkmark & 2,720 & 30,000 & 20,000              & 30,000            & 657,720 \\
v4 & --         & --         & --    & --     & 30,000 (1-site)     & 30,000$^{\dagger}$ & 60,000 \\
v5 & --         & --         & --    & --     & 40,000 (multi)      & 20,000$^{\dagger}$ & 60,000 \\
\bottomrule
\end{tabular}
\caption{Training-data composition by stage. v4 and v5 both resume from v3;
entries marked $^{\dagger}$ are anti-forgetting replay pairs drawn from the
v3 corpus.}
\label{tab:composition}
\end{table*}

\subsection{Inference}

At inference time both backbones generate corrections using beam search with
4 beams. We use \texttt{no\_repeat\_ngram\_size = 3} to prevent repetitive
output, length penalty 1.0, early stopping, and a maximum generation length
of 128 tokens. For mBART-50 we additionally set
\texttt{forced\_bos\_token\_id} to the \texttt{ta\_IN} language code to
ensure the decoder generates Tamil rather than another of the 50 languages
in the multilingual vocabulary. Each sentence requires a single forward
pass.

\subsection{Locality of Corrections}

Generating the corrected sentence in a single forward pass removes that
structural ceiling and yields one deterministic output per input, which
suits real-time use. The cost, as Section~\ref{sec:results} shows, is that
corrections are no longer localised: the model can also change words that
did not need changing.

\begin{table*}[t]
\centering
\small
\begin{tabular}{lccccc}
\toprule
 & mT5-small & mBART-v2 & mBART-v3 & mBART-v4 & mBART-v5 \\
\midrule
Parameters               & 300M & 610M & 610M & 610M & 610M \\
Resume from (per stage)  & --   & scratch & v2 & v3 & v3 \\
Epochs                   & 4 & 2 & 5 & 1 & 2 \\
Learning rate            & $3\times10^{-4}$ & $2\times10^{-5}$ & $5\times10^{-6}$ & $3\times10^{-6}$ & $8\times10^{-6}$ \\
Effective batch          & 128 & 128 & 128 & 128 & 128 \\
Optimizer                & Adafactor & AdamW & AdamW & AdamW & AdamW \\
Precision                & bf16 & bf16 & bf16 & bf16 & bf16 \\
Label smoothing          & -- & 0.1 & 0.1 & 0.1 & 0.1 \\
\bottomrule
\end{tabular}
\caption{Training hyperparameters per stage. All runs used an NVIDIA A100.}
\label{tab:hyper}
\end{table*}

\begin{table*}[t]
\centering
\small
\begin{tabular}{lccccccccc}
\toprule
 & \multicolumn{5}{c}{Fine-tuned seq2seq (this work)} & & \multicolumn{2}{c}{Tamil-LLaMA-7B} \\
\cmidrule(lr){2-6}\cmidrule(lr){8-9}
Category & mT5 v2 & mT5 v3 & mT5 v4 & mT5 v5 & mBART-50 v5 & Copy baseline & 0-shot & 3-shot \\
\midrule
Phonetic       & 50.5\% & 56.5\% & 54.5\% & 55.0\% & 69.5\% & 0.0\%   & 9.5\%  & 12.0\% \\
Subject--verb  & 1.0\%  & 52.5\% & 44.5\% & 40.0\% & 43.5\% & 0.0\%   & 44.5\% & 47.5\% \\
Pulli          & 63.0\% & 70.5\% & 69.0\% & 64.5\% & 72.0\% & 0.0\%   & 6.0\%  & 9.5\%  \\
Sandhi         & 0.0\%  & 0.0\%  & 63.0\% & 83.5\% & 87.5\% & 0.0\%   & 7.5\%  & 12.5\% \\
Identity       & 87.0\% & 82.0\% & 76.5\% & 72.5\% & 74.0\% & 100.0\% & 27.5\% & 42.0\% \\
\midrule
Overall        & 40.3\% & 52.3\% & 61.5\% & 63.1\% & 69.3\% & 20.0\%  & 19.0\% & 24.7\% \\
$\Delta$ vs copy & $+20.3$ & $+32.3$ & $+41.5$ & $+43.1$ & $+49.3$ & --- & $-1.0$ & $+4.7$ \\
\bottomrule
\end{tabular}
\caption{Per-category top-1 exact-match accuracy on the 1,000-sentence
balanced diagnostic set (200 per category). The copy baseline returns its
input unchanged, scoring 20.0\% by construction. Tamil-LLaMA is prompted,
not fine-tuned (Section~\ref{sec:llama}); its zero-shot result is not
statistically distinguishable from that baseline ($p = 0.59$).}
\label{tab:results}
\end{table*}

\section{Results and Discussion}
\label{sec:results}

We evaluate using top-1 exact-match accuracy at the sentence level on the
1,000-sentence balanced diagnostic set described in
Section~\ref{sec:diagnostic}, and additionally track token-F1 and character
error rate. Table~\ref{tab:results} reports per-category results for all
five checkpoints, the copy baseline and the two Tamil-LLaMA conditions.

\subsection{Headline Comparison}

mBART-50 v5 achieves the highest overall accuracy at 69.3\%, ahead of the
corresponding mT5-small stage (v5, 63.1\%) by 6.2 percentage points. The gap
is consistent with the architectural argument: mBART's denoising
pre-training objective --- reconstruct the original from a corrupted input
--- is essentially the task itself, whereas mT5's span-corruption objective
is a less direct match, and mBART's explicit language-ID conditioning is
better suited to a monolingual transformation than mT5's prefix-based task
signalling. mBART is also roughly twice the size, so the two effects are not
fully separable here.

\subsection{What Each Stage Contributes}

\paragraph{Contextual data is what teaches agreement.}
Subject--verb accuracy at v2 is 1.0\% --- effectively zero. The v2 model has
seen half a million surface-noise pairs and has learned to fix characters,
but it has no notion that \ta{அவள்} constrains the verb ending. Introducing
the contextual augmentation at v3 lifts this to 52.5\% (2/200 to 105/200)
without any change of architecture. v2 saw no agreement supervision at all,
so the jump reflects the introduction of the category rather than an
unusually large gain per training pair.

\paragraph{Cross-word sandhi requires cross-word supervision, and
multi-site supervision at that.}
Sandhi accuracy is 0.0\% at both v2 and v3, despite v3's 20,000 within-word
sandhi pairs. Only when genuinely cross-word pairs are introduced does the
capability appear: 63.0\% at v4 (single-site) and 83.5\% at v5
(multi-site). The 20.5-point v4--v5 gain comes purely from allowing 1, 2 or
3 sites per training sentence, teaching the model that a correction at one
position does not preclude another later on. mBART-50 v5 reaches 87.5\%
here, the strongest per-category result in the paper.

\subsection{The Sandhi--Identity Trade-off}

The most consistent pattern in Table~\ref{tab:results} is one that prior
Tamil spell-correction work does not report, because prior work does not
attempt sandhi: identity accuracy falls monotonically as sandhi accuracy
rises. Across the mT5 stages, identity moves $87.0 \rightarrow 82.0
\rightarrow 76.5 \rightarrow 72.5$ while sandhi moves $0 \rightarrow 0
\rightarrow 63.0 \rightarrow 83.5$. mBART-50 v5 shows the same relationship
at a better operating point (87.5\% sandhi at 74.0\% identity).

Inspection of the failures makes the mechanism clear: almost all identity
losses are the model applying sandhi to an already-acceptable sentence, for
example rewriting \ta{மாணவர்கள் அமைதியாக தேர்வு எழுதினார்கள்} as
\ta{மாணவர்கள் அமைதியாகத் தேர்வு எழுதினார்கள்}, or \ta{இந்த புத்தகம்} as
\ta{இந்தப் புத்தகம்}. These are not random corruptions: the suffixed variant
is prescriptively preferred while the unsuffixed one is widely used and was
labelled correct in our gold set. What the metric records as an identity
failure is therefore partly a disagreement about whether optional sandhi is
obligatory --- a contested question in Tamil prescriptive grammar. Since our
gold standard resolves it in one direction throughout, the identity and
sandhi columns measure conformity to a single convention, not adjudicated
ground truth.

For deployment the sandhi stages are therefore a tunable rather than a
strict improvement: a writing assistant that flags suggestions is well
served by v5's high sandhi recall, while a silent auto-correct is better
served by v3, which never touches a correct sentence for sandhi reasons.

\subsection{Per-Category Analysis}

\paragraph{Phonetic confusion.}
mBART-50 v5 reaches 69.5\%, ahead of every mT5 stage (50.5--56.5\%). Common
confusions (\ta{ழ}$\leftrightarrow$\ta{ல}, \ta{ள}$\leftrightarrow$\ta{ல})
are handled reliably by both backbones; residual failures concentrate on
rarer pairs and word-initial positions, where less context is available.

\citet{yazhmozhi2026tamilmayangoli} report 93.50\% exact match on the same
confusion groups, against our 69.5\%. The figures are not on the same scale:
their models are trained and tested on that one class alone, with a single
dictionary-constrained substitution per sentence drawn from the same induced
distribution as training, whereas ours must select among ten categories on a
separately constructed set and leave 200 sentences untouched. Their
cross-genre scores match in-domain validation exactly, which they attribute
to controlled induction flattening genre differences --- a caveat our
Wikipedia bias shares.

\paragraph{Subject--verb agreement.}
This is the one category where the smaller model wins: mT5 v3 achieves
52.5\% against mBART-50 v5's 43.5\%. The mT5 trajectory also declines after
v3 ($52.5 \rightarrow 44.5 \rightarrow 40.0$), which indicates that the
sandhi-focused stages induce partial forgetting of agreement despite the
replay sample. This is a concrete, actionable finding: the replay fraction
for v4 and v5 is currently drawn uniformly from the v3 corpus, and weighting
it toward agreement pairs is the obvious next experiment.

\paragraph{Pulli omission.}
Performance is stable in the 63--72\% band, best at 72.0\% (mBART-50 v5).
Pulli restoration is a local decision, and neither the contextual nor the
sandhi stages change it much.

\subsection{Comparison with Tamil-LLaMA}
\label{sec:llama}

We evaluate Tamil-LLaMA-7B-Instruct \citep{balachandran2023tamilllama} on
the same 1,000-sentence diagnostic set in 4-bit NF4 quantization, using the
model's native Alpaca format with English and Tamil instructions, zero-shot
and with three demonstrations drawn from the training pool and therefore
disjoint from the diagnostic set by construction. The best-scoring prompt of
each kind is selected on a stratified 100-sentence probe; scores spanned
20.0--25.0\%, within noise at $n = 100$, and excluding probe items moves the
reported figures by 0.2 points or less.

Zero-shot Tamil-LLaMA reaches 19.0\% (95\% Wilson CI 16.7--21.5), which an
exact McNemar test cannot distinguish from the 20.0\% copy baseline (135
won, 145 lost, $p = 0.59$). Three-shot prompting reaches 24.7\% (95\% CI
22.1--27.5), a significant but small gain over both the copy baseline
($p = 0.006$) and zero-shot ($p = 1.3\times10^{-7}$), against 69.3\% for
mBART-50 v5.

Few-shot prompting buys restraint rather than skill: demonstrations raise
identity accuracy from 27.5\% to 42.0\% and cut the rate at which the model
edits an already-correct sentence from 144/200 to 115/200, while errors left
uncorrected rise from 134/800 to 258/800. The model is competitive only on
subject--verb agreement (47.5\%), where its pre-training prior over Tamil
verb morphology applies directly. On pulli and sandhi it produces the
required edit in 21.0\% and 17.0\% of cases but reaches exact match on only
9.5\% and 12.5\%, because it simultaneously rewrites unrelated parts of the
sentence. Prompting alone therefore does not close the gap.

\subsection{Key Findings}

\begin{enumerate}
\item \textbf{Curriculum matters more than any single dataset.} Each
  category became learnable only when supervision of exactly that kind was
  introduced --- agreement at v3, cross-word sandhi at v4, multi-site sandhi
  at v5. Neither more surface noise nor a larger backbone substituted for
  the right data.
\item \textbf{Gains in one category are not free.} Sandhi recall is bought
  with identity precision, monotonically --- an aggregate number would have
  concealed this, which is the argument for category-balanced evaluation.
\item \textbf{Prompting does not substitute for task supervision.}
  Tamil-LLaMA-7B scores 19.0\% zero-shot and 24.7\% few-shot against a
  20.0\% copy baseline, the zero-shot condition statistically inseparable
  from it. Its accuracy concentrates in the one category where a
  language-model prior transfers directly, and its dominant failure mode is
  rewriting text that needed no rewriting --- a characterisation of prompted
  transfer to a narrow orthographic task, not of the generation the model
  was built for.
\end{enumerate}

\section{Conclusion}

We present a context-aware Tamil spell and grammar correction system based
on end-to-end fine-tuning of pre-trained multilingual seq2seq transformers
under a four-stage progressive schedule. Our best model, mBART-50 v5
(610M), reaches 69.3\% top-1 exact-match accuracy on a 1,000-sentence
balanced diagnostic set verified disjoint from training data, with 87.5\% on
cross-word sandhi --- a category no prior Tamil spell checker attempts. The
staged ablation shows that subject--verb agreement becomes learnable only
with contextual supervision (1.0\% $\rightarrow$ 52.5\%) and cross-word
sandhi only with multi-site cross-word supervision (0\% $\rightarrow$
83.5\%), and it exposes a systematic trade-off in which sandhi recall is
paid for in identity precision. Tamil-LLaMA-7B-Instruct, prompted on the
same test set, trails mBART-50 v5 by 44.6 points, indicating that
task-specific supervision, not model scale, is what this problem currently
requires.

\paragraph{Future scope.}
The most direct extensions are weighting the v4/v5 replay sample toward
agreement pairs, and extending the multi-site sandhi dataset to the contexts
v5 still misses. Beyond that, LoRA fine-tuning of Tamil-LLaMA on the same
data would separate task supervision from architecture; training on a
genre-balanced corpus such as TamilCorp \citep{yazhmozhi2025building} would
address the Wikipedia bias directly, and adding NLLB would test the
curriculum beyond the mT5/mBART pair; a confidence threshold on optional
sandhi would make the sandhi--identity trade-off a deployment parameter;
adjudication by native Tamil speakers would settle the optional-sandhi
convention; and authentic errors from social media and student writing would
test generalisation beyond Wikipedia noise.

\section*{Limitations}

\paragraph{Partial sandhi coverage.}
Sandhi failures cluster in specific syntactic patterns, notably
dative-marked nouns followed by certain verbs, which the multi-site
generator undersamples relative to their difficulty. Rarer phenomena such as
nasal assimilation are absent entirely, so the reported 87.5\% covers
\emph{vallinam} sites alone.

\bibliography{custom}

\end{document}